# Neural Contrastive Clustering: Fully Unsupervised Bias Reduction for Sentiment Classification

Jared Mowery

## Abstract

**Background:** Neural networks produce biased classification results due to correlation bias (they learn correlations between their inputs and outputs to classify samples, even when those correlations do not represent cause-and-effect relationships).

**Objective:** This study introduces a fully unsupervised method of mitigating correlation bias, demonstrated with sentiment classification on COVID-19 social media data.

**Methods:** Correlation bias in sentiment classification often arises in conversations about controversial topics. Therefore, this study uses adversarial learning to contrast clusters based on sentiment classification labels, with clusters produced by unsupervised topic modeling. This discourages the neural network from learning topic-related features that produce biased classification results.

**Results:** Compared to a baseline classifier, neural contrastive clustering approximately doubles accuracy on bias-prone sentences for human-labeled COVID-19 social media data, without adversely affecting the classifier's overall F1 score. Despite being a fully unsupervised approach, neural contrastive clustering achieves a larger improvement in accuracy on bias-prone sentences than a supervised masking approach.

**Conclusions:** Neural contrastive clustering reduces correlation bias in sentiment text classification. Further research is needed to explore generalizing this technique to other neural network architectures and application domains.

**Correspondence:** jared.mowery@gmail.com

# Introduction

Crises such as the COVID-19 pandemic can create rapidly emerging and unpredictable sources of bias in sentiment classification, and effectively mitigating these biases requires an automatic and fully unsupervised approach. Currently, most bias mitigation approaches are only effective in supervised or weakly supervised settings, and in both of these settings, the approaches are typically limited to mitigating demographics-related sources of bias, such as gender bias. Unknown, unpredictable sources of bias which are specific to a new subject, like the COVID-19 pandemic, are harder to mitigate since it is difficult to anticipate which sources of bias will emerge, and it is time-consuming to produce supervised data to retrain classifiers to address each new source of bias. This study addresses these challenges by providing a fully unsupervised approach for mitigating correlation bias, which occurs when classifiers learn to perform classification tasks using correlations between words and emotions, even when those correlations do not represent expressions of emotion. For example, words like "hydroxychloroquine", "virus", and "mandate" may not normally correlate with an emotion in most situations, but during the COVID-19 pandemic, these words become frequently correlated with expressions of anger in online discussions, even though the words themselves convey no emotion. Without correlation bias mitigation, the classifier learns these correlations, and consequently becomes biased to interpret sentences containing these words as expressing anger.

A recent survey [1] of bias mitigation approaches, which described multiple types of bias and the supervised bias mitigation algorithms that have been developed so far, highlighted the need for unsupervised bias mitigation algorithms. The survey authors found three unsupervised bias mitigation studies, but none of the three studies were designed to mitigate correlation bias, although one may indirectly reduce correlation bias.

This paper introduces a fully unsupervised bias mitigation algorithm for correlation bias, and demonstrates its effectiveness in sentiment classification for COVID-19 social media posts.

The remainder of this section will provide a brief overview of bias and discrimination in neural networks and bias mitigation research. Since unsupervised techniques for bias mitigation are rare and have only been developed for certain types of bias, this overview will also include other techniques which may be adapted to unsupervised bias mitigation. This includes adversarial learning in supervised bias mitigation techniques, weakly supervised techniques, and several computer vision studies.

Bias and discrimination in neural networks raise numerous cross-disciplinary questions involving legal, ethical, and societal factors [2]. Bias exists in many natural language processing technologies, including fundamental technologies for sentiment classification: word embeddings, language models, and the sentiment classifiers themselves. Word embeddings have been proven to contain human-recognizable bias by demonstrating a statistically significant correlation between implicit association tests used by psychologists and cosine similarity distances between baskets of relevant words in the word embedding [3], and a similar result holds for image classification [4]. A survey of bias in natural language generation reveals considerable challenges [5], and language models have specifically been shown to exhibit bias in expressed sentiment when generating text based on country names, occupations, and genders [6]. Systematic race and gender bias has been detected in an analysis of over 200 sentiment analysis systems [7]. The evidence for bias in sentiment classification systems raises significant risks for real world applications, especially when the results may be used to inform policy makers or in automated processing systems that could exhibit discrimination.

To mitigate this bias, supervised and weakly supervised techniques have been developed which leverage lists of protected attributes, correlated attributes or word lists to mitigate bias.

Supervised techniques have been developed to mitigate bias when protected attributes are known and data labeled with those attributes is available, especially when the protected attributes are demographic, such as age, gender, and race [8]. Adversarial representation learning has been applied with known protected attributes in conjunction with Kullback-Leibler divergences between output distributions and desired, unbiased distributions to minimize bias in the network (e.g. in computer vision [9]), or in adversarial learning approaches in which part of a classifier learns to perform the intended classification task while another part "unlearns" how to predict the protected attribute (e.g. [10] [11] [12] [13] [14]). In cases where protected attributes are unavailable but correlate with available attributes, such as detected dialect correlating with race [15] or race correlating with ZIP codes, the correlated attributes can be used as proxies for the protected attributes [16] [17].

For some protected attributes, weakly supervised techniques have been developed to alleviate the need for labeled data. These approaches often rely on word lists that refer to a protected attribute, such as gendered words or references to race. Examples include reducing gender bias in word embeddings by augmenting the text data with versions in which gendered words are swapped [18], minimizing the differences between gendered words on a subspace representing gender bias through adversarial learning [19], introducing a loss term to minimize a projection of an encoder-trained word embedding onto a subspace that encodes gender [20], and leveraging pairs of word lists that indirectly define biases [21]. Word lists that have been manually labeled to indicate whether each word is relevant to a classification task or bias-prone can also help reduce bias. These lists can be used to reduce the bias neural networks learn from their training data via objective functions that encourage learning features from relevant words and discourage learning features related to bias-prone words [22], or to mask input words for a trained neural network to prevent it from using bias-prone words in its sentiment classification decisions [23].

Generating lists of protected attributes, correlated attributes or word lists is labor intensive, so some semi-automated techniques have been developed to reduce the labor required to generate these lists of relevant or bias-prone tokens (e.g. words, URLs, hashtags, names), as well as help researchers explore types of bias. For example, Ferrer et al. [24] [25] use a pair of hand-chosen sets of words representing a type of bias (e.g. lists of male and female gendered words to represent gender bias) to define a type of bias to explore. They automatically identify candidate words that may be expressions of bias related to those word lists by calculating the centroid of each word list in a word embedding, and estimating the bias-propensity of each remaining word based on its frequency of occurrence in the corpus and proximity to either centroid. Each word is scored for positive or negative sentiment using pre-trained algorithms, and words belonging to parts of speech that are unlikely to exhibit bias (e.g. articles and proper nouns) are ignored. Finally, clusters of words that exhibit strong negative sentiment and a close proximity to either centroid can be clustered and categorized using a semantic analysis system to give an end user clues about the nature of the bias in a given corpus of text, such as showing that many of the candidate gender biased terms used by a Reddit community are associated with the semantic categories of "Relationship: Intimate/Sexual" and "Power, Organizing". Approaches like these reduce human annotation labor by requiring only the initial word lists to estimate the likelihood that other words may reflect bias.

The previously mentioned bias mitigation survey that highlighted a lack of unsupervised algorithms [1] cited three papers covering two types of unsupervised algorithms. The first type of unsupervised algorithm compensates for bias arising from the distribution of the training data: sparse portions of the input distribution may correspond to underrepresented groups and achieve lower classification accuracy compared to dense portions of the input space, so these algorithms use distributionally robust optimization [26] [27] or adversarial weighting of samples [28] to ensure fairer accuracy across the input space. The second type of unsupervised algorithm develops disentangled feature representations for a computer vision task and demonstrates a correlation between using the disentangled representations and fairness [29]. Disentangling feature representations may produce features that are less likely to conflate protected attributes and features relevant to the classification task. This could increase the odds that classifiers focus on the relevant features, and consequently indirectly address correlation bias. However, a more direct approach is still needed to mitigate correlation bias, since the disentangled features for protected attributes will often correlate with the classification labels, and consequently be learned by classifiers. Future research could explore combining disentangled features with correlation bias mitigation to improve the degree of correlation bias mitigation.

The algorithm most closely related to this study is a fully unsupervised, hybrid statistical and machine learning technique that discovers bias-prone tokens and mitigates bias in sentiment classifiers, without retraining the classifier [30]. It identifies bias-prone words by contrasting clusters of sentences labeled as expressing anger or negative sentiment against clusters of sentences produced via unsupervised topic modeling. Since many bias-prone words pertain to controversial topics, words that are more strongly associated with topics are disproportionately likely to be bias-prone (e.g. "hydroxychloroquine"), while words that are more strongly associated with the emotion or sentiment labels (e.g. "hate") are more likely to be expressions of emotion or sentiment. The algorithm uses a pre-trained word embedding to improve accuracy in its classification of words as bias-prone by recognizing that bias-prone words' nearest neighbors in the embedding space should not be expressions of emotion or sentiment, and vice versa. The algorithm's identification of bias-prone words is sufficiently accurate that automatically masking them in existing sentiment and emotion classifiers' inputs mitigates bias, without the algorithm requiring a specification of types of bias to mitigate, protected attributes, or hand-chosen word lists. The algorithm presented in this study also uses unsupervised topic modeling to achieve unsupervised bias mitigation, but takes a different approach: this study uses adversarial learning to debias the neural network itself and eliminate the need for masking, rather than masking inputs to an already trained neural network.

While the scope of this paper is limited to *correlation* bias, the lack of unsupervised algorithms for mitigating any forms of bias makes existing unsupervised algorithms noteworthy, in addition to the unsupervised correlation bias mitigation algorithm described in the previous paragraph.

## Methods

This study presents a fully unsupervised algorithm for mitigating correlation bias via adversarial training and unsupervised topic modeling. Since topic modeling relies on word embeddings, which in turn reflect human biases [3], a classifier can be debiased by "unlearning" topic classification while learning to classify sentiment. However, while [30] treated the sentiment classifier as a black box and masked its inputs to achieve a reduction in bias, this study incorporates topic classification into the training process via adversarial learning. As bias mitigation research continues, in theory, de-biasing the network itself should outperform masking, since masking ignores the context of words, while the network can unlearn bias in a context-aware manner and also unlearn bias from words indirectly associated with the bias-prone words.

The remainder of this section will describe preparing the data (Data Preparation) and the adversarial learning algorithm (Adversarial Learning).

**Data Preparation**

Reddit social media discussion forum data was gathered to produce training, development, and test data sets using a two step query process. The first step queried for posts containing COVID-19 specific terms, such as "COVID-19" and "coronavirus". The second step segmented those posts into sentences and kept only sentences matching a much broader query for pandemic subtopics (e.g. "vaccine" or "CDC"). This two-step process is designed to gather a wide diversity of data: the first step ensures posts are relevant to COVID-19, while the second step ensures sentences are included across a wide range of pandemic subtopics. If only the query terms from the first step were used, the resulting data set would be too narrowly focused on words like "COVID-19", which would not be representative of online discussions and could produce an artificially simple sentiment classification problem.

Many sentiment analysis studies use weakly supervised query terms (e.g. emojis or hashtags) or explicit sentiment-bearing terms (e.g. "angry") to gather posts that contain a higher likelihood of expressing sentiment. While this approach reduces the number of sentences that need to be annotated to obtain a reasonable number of sentences expressing anger, it would jeopardize the validity of this study since it can also introduce biases into the training data: the data set would include an artificially narrow or homogeneous subset of expressions of sentiment, and those expressions of sentiment could be disproportionately overt or susceptible to bias. Therefore, no sentiment-related terms were used in either querying step.

The posts were gathered from December 2019 through November 2021. A random sample of 20,043 sentences were then annotated for whether they (1) express anger, and (2) could easily be misinterpreted as an expression of anger *by a biased annotator*. For brevity, *bias-prone* sentences will be used to refer to the sentences that could be labeled as expressing anger by a biased annotator. Anger was chosen due to its prevalence in the data and the overall rarity of expressions of sentiment in the data (due to not using sentiment-specific query terms). The annotation process yielded 263 bias-prone sentences (1.3%) and 3,477 angry sentences (17.3%). However, note that the 1.3% prevalence of bias-prone sentences is probably an underestimate, since it reflects sentences with relatively straightforward biased interpretations, and that the misclassification rate in sentiment classifiers due to correlation bias is likely to be much higher. The annotations were split into training (60%), development (20%), and test (20%) data sets. Non-angry sentences were deleted from the training data set at random to equalize the number of angry and non-angry sentences, reducing the number of sentences from 12,025 to 4,172, split evenly between angry and non-angry labels.

To compare the algorithm in this study to an idealized case of masking bias-prone tokens (i.e.

replacing those tokens with "it" so they do not influence the sentiment classifier's decisions), the annotator also labeled the 10,000 tokens that occur most frequently in the corpus to indicate whether they represent expressions of sentiment, and whether they belong to topics that can be the subject of angry discussions (and, therefore, likely sources of correlation bias). The annotator labeled 373 as expressions of sentiment and 1,029 as potentially bias-prone.

Finally, the sentences were given topic labels using BERTopic [31], a topic modeling algorithm. To reduce the topic labels to a binary classification problem, a sentence was labeled as having a recognizable topic if the maximum-scoring topic from BERTopic was greater than 0.25 (on a zero to one scale), and that score was greater than a score BERTopic provides for a miscellaneous topic that indicates the sentence had no recognizable topic. BERTopic was initialized to find 50 topics using 113,438 unlabeled sentences from across the December 2019 to November 2021 time period, and then used to label the annotated data with topics. The 50 topics included the miscellaneous topic and provided a reasonable variety of topics.

**Adversarial Learning**

This study uses a three-part neural network architecture consisting of a shared component and two classification heads. The shared component includes the pre-trained RoBERTa language model from TweetEval [32] coupled with a maximum pooling layer or a Gated Recurrent Unit (GRU) [33] recurrent neural network layer followed by a maximum pooling layer. Both classification heads use a pair of fully connected layers, with one classification head dedicated to sentiment classification and the other to topic classification. Dropout regularization [34] is used in all three components of the neural network.

The adversarial learning approach trains the shared component and sentiment classification head to maximize sentiment classification accuracy. Meanwhile, it trains the topic classification head to maximize topic classification accuracy and trains the shared component to *minimize* topic classification accuracy, via inverted topic labels. To formalize this concept, define the trainable parameter sets $\theta$ for the shared component (language model and GRU), $\phi$ for the topic classification head, and $\psi$ for the sentiment classification head. Define the training data sets as $D_s$ for sentiment classification, $D_t$ for topic classification, and $D_{-t}$ for "anti-topic" classification, in which the positive and negative class labels in $D_t$ have been swapped. Next, let $x$ denote the input, $z$ the encoded latent representation (the output of the shared component), $\hat{y}_s$ the predicted sentiment labels, $\hat{y}_t$ the predicted topic labels, and $\hat{y}_{-t}$ the predicted anti-topic labels. Letting $L$ represent the binary cross-entropy loss function, the neural network's objective function is:

$$max_\theta\, max_\phi\, max_\psi\, L_{\hat{y}_s}(D_s;\theta,\psi) + L_{\hat{y}_t}(D_t;\phi) + L_{\hat{y}_{-t}}(D_{-t};\theta)$$

Define the encoders and decoders as $f_\theta : x \rightarrow z$ for the encoder, $f_\psi : z \rightarrow \hat{y}_s$ for the sentiment decoder, $f_\phi : z \rightarrow \hat{y}_t$ for the topic decoder, and $f_\theta : x \rightarrow z$ for the anti-topic decoder. Now the three-step optimization procedure for the objective function can be written as:

1. Sentiment training: train $f_\theta$ and $f_\psi$ on $D_s$ and $y_t$ while holding $\phi$ fixed
2. Topic training: train $f_\phi$ on $D_t$ while holding $\theta$ and $\psi$ fixed
3. Anti-topic training: train $f_\theta$ (the encoder) on $D_{-t}$ while holding $\phi$ and $\psi$ fixed

The data sets $D_t$ and $D_{-t}$ use the same samples for each batch of training data and the order of the samples is jointly randomized for each training epoch, so steps 2 and 3 will always perform regular and adversarial topic classification updates on the same sample sentences. The weight updates for step 2 are completed prior to any calculations for step 3, to avoid creating potential discontinuous weight updates at the junction between the shared model components and the topic classification head. $D_s$ uses its own samples for each training epoch and the order is randomized independently of $D_t$ and $D_{-t}$, to permit the sentiment and topic data sets to be independently balanced and to reduce the risk of introducing instabilities.

The training procedure also includes an optional loss coefficient for the topic and anti-topic loss functions. The loss coefficient controls the relative importance of topic/anti-topic and sentiment classification accuracy (the sentiment loss function's coefficient is always 1). The topic loss coefficient can be used to control the relative importance the algorithm places on debiasing versus sentiment classification accuracy, and consequently, on which features the classifier learns based on how strongly each feature correlates with sentiment labels versus topic labels. In current adversarial debiasing, the protected attributes are known and assumed to be the complete set of protected attributes, so it is intuitive to completely eliminate the classifier's ability to predict the protected attributes. However, in this study, the protected attributes are unknown, and only hypothesized to correlate with topics. Therefore, it is not necessarily desirable to eliminate all topic-related features from the latent representation $z$, and the topic coefficient provides a mechanism to determine the extent to which topic-related features are removed from $z$.

The topic training data, $D_t$, consists of 4,172 sentences that were sampled from the annotated training data, with a 50/50 split between sentences containing topics and sentences belonging to the miscellaneous topic (i.e. the sentence had no recognizable topic). The number was chosen to exactly equal the size of the sentiment training data. However, to potentially reduce the risk of instability during adversarial training, the subset of annotated sentences used to produce the topic data set was chosen independently from the 12,025 sentiment training sentences (i.e. prior to balancing the number of angry and non-angry sentences), which should reduce the risk of multiple, significant weight updates affecting the same set of parameters during batch training. The sentences for the miscellaneous topic were chosen at random.

Since correlation bias arises from correlations between contentious topics and emotion labels, the bias mitigation effectiveness of the topic and anti-topic training can be increased by ensuring the topic training data set includes the topics which most frequently co-occur with expressions of anger. The remainder of this section describes the sampling strategy used to select the 2,086 sentences with recognizable topics for $D_t$.

To define a metric for the topic and anger co-occurrence frequency, let each topic $t$ in the annotated training data, excluding the miscellaneous topic, consist of $|t|$ sentences, and let the set of training data sentences labeled with anger be $A$. Now define the lower bound of a 95% confidence interval for the probability that a sentence $s$ in $t$ is labeled with anger as:

$$\widetilde{P}(s \in A | s \in t) = Wilson(|t \cap A|, |t|)$$

where *Wilson* is the lower bound of the Wilson confidence interval with continuity correction [35]. Using the lower bound of a confidence interval accounts for sample size, so that topics with very few sentences do not spuriously appear to be highly correlated with anger. Define the expected number of sentences in *t* expressing anger as:

$$n_t = \widetilde{P}(s \in A | s \in t) \cdot |t|$$

Next, normalize the set of expected values across all topics *T* by dividing by their sum:

$$\bar{n}_t = \frac{n_t}{\sum_{i \in T} n_i}$$

Finally, the number of sentences to sample for each topic *t* is $2{,}086 \cdot \bar{n}_t$. This ensures $D_t$ contains sentences representative of the topics that most frequently overlap with expressions of anger, consequently improving the effectiveness of the adversarial training procedure in mitigating correlation bias.

## Results

This section quantifies the efficacy of the unsupervised correlation bias mitigation method by testing eight main classifier variations for their overall F1 scores and for the fraction of bias-prone sentences that they correctly mark as not expressing anger. The eight classifier variations are defined by three parameters: whether the classifier is regular (R) or debiased (D), whether supervised masking is applied to the test data to help reduce bias, and whether the classifier includes the GRU layer or only a maximum pooling layer. Masking was implemented by replacing bias-prone tokens with "it", using the annotated list of 1,029 bias-prone tokens. This represents an idealized masking approach, since the annotations were produced by hand rather than an unsupervised algorithm.

The results demonstrate the efficacy of unsupervised correlation bias mitigation (Table 1). The first four rows represent the typical, unmasked versions of the classifiers, with the debiased version improving the bias accuracy by 0.34 (without GRU) or 0.29 (with GRU) points compared to the regular classifier. The next four rows show that debiasing outperforms masking, and debiasing alone outperforms a combination of debiasing and masking.

Due to the small number of hand-annotated examples of bias-prone sentences, the results were averaged across epochs 3 through 9, inclusive. All classifiers had converged by epoch 3, and none exhibited signs of over-fitting through epoch 9, so averaging over those 7 epochs helps to reduce variance in the test results due to fluctuations in the neural network's trainable weights. In addition, for the less computationally expensive non-GRU classifier, the test was repeated five times and the results were averaged.

Table 1: Comparison of regular (R) and debiased (D) classifiers, with and without Gated Recurrent Unit (GRU) layers and masking, for 4,009 test sentences. Results include the overall F1 score on the test data and the accuracy on a subset of 263 bias-prone sentences. Without masking, debiasing improves the bias accuracy by 0.34 points without a GRU and by 0.29 points with a GRU. Debiasing outperforms both supervised masking alone and a combination of masking and debiasing.

| Classifier | GRU | Mask | F1 | Bias Accuracy |
|---|---|---|---|---|
| R | N | N | 0.61 | 0.33 |
| R | Y | N | 0.6 | 0.31 |
| D | N | N | 0.59 | **0.67** |
| D | Y | N | 0.63 | **0.6** |
| R | N | Y | 0.59 | 0.42 |
| R | Y | Y | 0.59 | 0.43 |
| D | N | Y | 0.58 | **0.57** |
| D | Y | Y | 0.6 | **0.54** |

## Discussion

The results demonstrate that accuracy on bias-prone sentences can be approximately doubled for sentiment classification, even using a fully unsupervised approach. Unsupervised debiasing also outperformed an idealized debiasing approach using hand-annotated word lists to mask words. Further research is needed to explore variations on this approach that may improve its effectiveness, to develop heuristics or algorithmic methods for selecting the correct topic loss coefficient (all experiments used loss coefficients of 1), and to generalize this approach to other application domains.

In terms of generalization, this study relies on a simple premise: correlation versus causation can be approximated by contrasting the utility of features for the intended classification task against their utility in classifying samples according to an unsupervised clustering algorithm. While this could be applicable to a wide variety of classification problems, it may be especially well-suited to natural language processing. Languages evolved to permit (reasonably) clear communication, which makes ambiguity undesirable. This encourages words to have relatively few semantically distinct meanings, especially in context and for their most commonly used meanings. As a result, discovering that a contextualized word or phrase has a topic-related meaning makes it unlikely that it will have a sentiment-related meaning, and vice versa. In application domains outside natural language processing, this property may be weaker, resulting in the debiasing technique producing weaker results. Future research could explore combining the algorithm presented in this paper with disentangling features to improve accuracy.

In terms of limitations, the same simple premise may sometimes cause a classifier to unlearn useful features, by mistaking them for features that only correlate with classification labels. This cannot be directly addressed until machine learning algorithms are capable of symbol grounding (learning what input features mean beyond a latent semantic sense) and reasoning over plausible cause-and-effect relationships. In the meantime, future research could investigate using the topic loss coefficients to control the trade-off between retaining useful features and rejecting features that only correlate with the classification task. Extensions to the selection and weighting of topics could also be explored, including replacing the existing binary topic classification heads with multilabel heads to better account for individual variation in topics, especially when certain topics may be especially related to the classification problem. For example, references to death and destruction will usually correlate with both negative sentiment and violent topics, leading to bias. In contrast, for discussions of fictional death and destruction, such as murder mysteries and video games, certain expressions of death and destruction may be expressions of positive sentiments, leading to the opposite bias. In such cases, it may be necessary to choose the topics and topic weights used as a basis for debiasing carefully, depending on whether the subject matter for the classifier is known in advance or if the goal is to build a general purpose classifier (e.g. by including or excluding entertainment related topics). Topic modeling over time may also be an interesting direction for future research, since sources of bias often change. For example, "hydroxychloroquine" may have initially been correlated with positive sentiment before switching to a correlation with negative sentiment. More broadly, this study implements a heuristic for distinguishing correlation from causation, and future research may yield improved heuristics.

The results revealed that combining unsupervised debiasing with masking actually yielded poorer accuracy on bias-prone sentences than debiasing alone. This may indicate debiasing discourages the classifier from learning obvious, bias-prone terms during training. Since those same terms will often be in the set of masked terms, debiasing may reduce the effectiveness of masking. Exploring masking techniques that identify more subtle bias-prone features (e.g. terms or combinations of terms) may still yield an improvement when combined with debiasing, and provide a method for measuring how well the adversarial training procedure identifies and removes these more subtle bias-prone features.

Finally, the neural network architecture may significantly impact the effectiveness of the debiasing technique. For example, some network architectures use latent feature representations with less representational power, such as using a lower dimensionality $z$ or using a separator token's embedding representation instead of maximum pooling. These architectures may comingle relevant features with bias-prone features to an extent that the debiasing algorithm cannot unlearn the bias-prone features while still achieving reasonable classification accuracy.

## Conclusion

Neural contrastive clustering is a fully unsupervised approach that approximately doubles accuracy in sentiment classification of bias-prone sentences for COVID-19 data, and outperforms supervised masking. Further research could explore improvements and generalizations of the neural contrastive clustering technique, as well as investigating its links to other unsupervised bias reduction methods for other types of bias, such as distributionally robust optimization for input distribution bias. Combining neural contrastive clustering with disentangled feature representations developed for computer vision tasks may also improve its effectiveness.